\pdfoutput=1

\documentclass[11pt]{article}

\usepackage[final]{acl}

\usepackage{times}
\usepackage{latexsym}

\usepackage[T1]{fontenc}

\usepackage[utf8]{inputenc}

\usepackage{microtype}

\usepackage{inconsolata}

\usepackage{url}
\usepackage{graphicx}
\usepackage{subcaption}
\usepackage{booktabs}
\usepackage{multirow}

\usepackage{amsmath}

\newcommand{\mc}[3]{\multicolumn{#1}{#2}{#3}}
\newcommand{\hy}[1]{\textcolor{black}{#1}}
\newcommand{\Ni}{({\em i})~}
\newcommand{\Nii}{({\em ii})~}
\newcommand{\Niii}{({\em iii})~}

\title{Sign Language Translation with Sentence Embedding Supervision}

\author{Yasser Hamidullah \and Josef van Genabith \and Cristina España-Bonet \\ { \tt \{yasser.hamidullah,Josef.van{\_}Genabith,cristinae\}@dfki.de} \\German Research Center for Artificial Intelligence (DFKI GmbH)\\
Saarland Informatics Campus, Saarbrücken, Germany
}

\begin{document}
\maketitle
\begin{abstract}
State-of-the-art sign language translation (SLT) systems facilitate the learning process through gloss annotations, either in an end2end manner or by involving an intermediate step.
Unfortunately, gloss labelled sign language data is usually not available at scale and, when available, gloss annotations widely differ from dataset to dataset. We present a novel approach using sentence embeddings of the target sentences at training time that take the role of glosses.
The new kind of supervision does not need any manual annotation but it is learned on raw textual data.
As our approach easily facilitates multilinguality, we evaluate it on datasets covering German (PHOENIX-2014T) and American (How2Sign) sign languages and experiment with mono- and multilingual sentence embeddings and translation systems. 
Our approach significantly outperforms other gloss-free approaches, setting the new state-of-the-art for data sets where glosses are not available and when no additional SLT datasets are used for pretraining, diminishing the gap between gloss-free and gloss-dependent systems.

\end{abstract}

\section{Introduction}
\label{s:intro}

Sign Language Translation (SLT) aims at generating text from sign language videos. There are several approaches to SLT reported in the literature, with \emph{sign2text} and \emph{sign2gloss2text} the most widely used.
While sign2text directly translates video into text with or without the help of glosses~\citep{camgoz18}, sign2gloss2text passes through an intermediate gloss step before translation into spoken language text~\citep{ormelrec}.
That is, sign2gloss2text breaks down the problem into two independent sub-problems using \textit{glosses} as a pivot language. A gloss is a textual label associated with a sign, and, although human signers do not in general use them, performance in automatic SLT has long been upper bounded by the gloss supervision and their use as an intermediate representation \citep{camgoz18}.
The advantage of translation without glosses is that collecting data is much easier. Even though translation results are better for approaches that use glosses as intermediate representation \citep{chenEtAl:2022,chen2022two}, this comes at the cost of annotating all the video data with glosses which is a time consuming manual task. For many data sets glosses are simply not available.
On the plus side, with gloss supervision-based SLT architectures, one can take full advantage of the maturity of text2text machine translation between glosses and spoken language text.

In this work, we present a novel approach \emph{sign2(sem+text)}, a model that gets rid of glosses and adds supervision through sentence embeddings, SEM, pretrained on raw text and finetuned for sign language.
%
Our experimental results demonstrate the strength of the novel approach on both standard small datasets with gloss annotation and larger datasets without. In the latter case, we achieve state-of-the-art results for the American Sign Language (ASL) dataset How2Sign when no additional SLT datasets are used%
\footnote{\citet{UthusSLhw1}  and \citet{rust2024privacyaware} obtain better results by using the YouTube-ASL dataset for pretraining.}
 improving  over \citet{Tarres_2023_CVPR} by 4 BLEU points. For German Sign Language (DGS), our new approach achieves translation quality scores between the previous best gloss-free system \citep{Zhou_2023_ICCV} and the current state-of-the-art using glosses \citep{chen2022two} on the PHOENIX-2014T dataset.
Our code and models are publicly available.%
\footnote{\url{https://github.com/yhamidullah/sem-slt}}




\begin{figure*}[htbp]
        \includegraphics[width=0.46\textwidth]{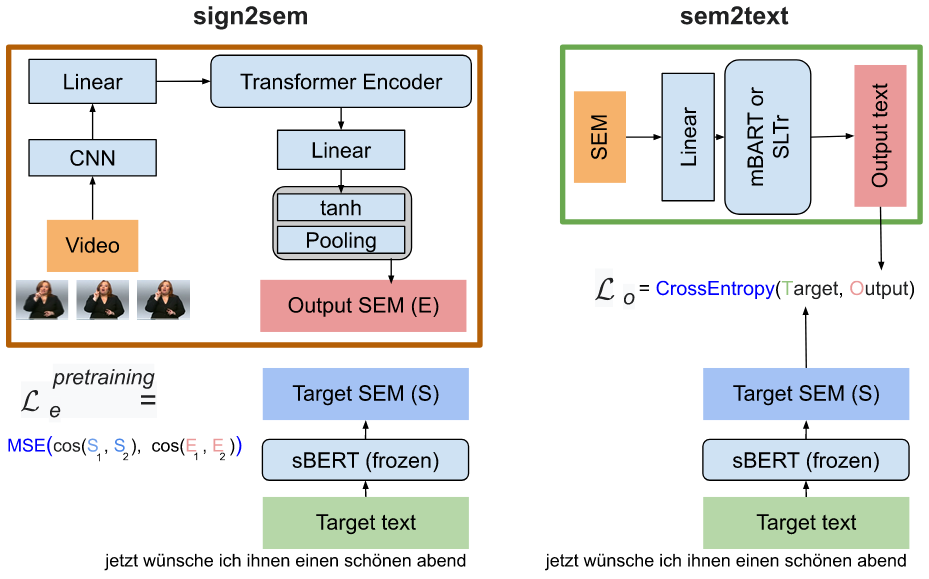}
       \hfill
        \includegraphics[width=0.44\textwidth]{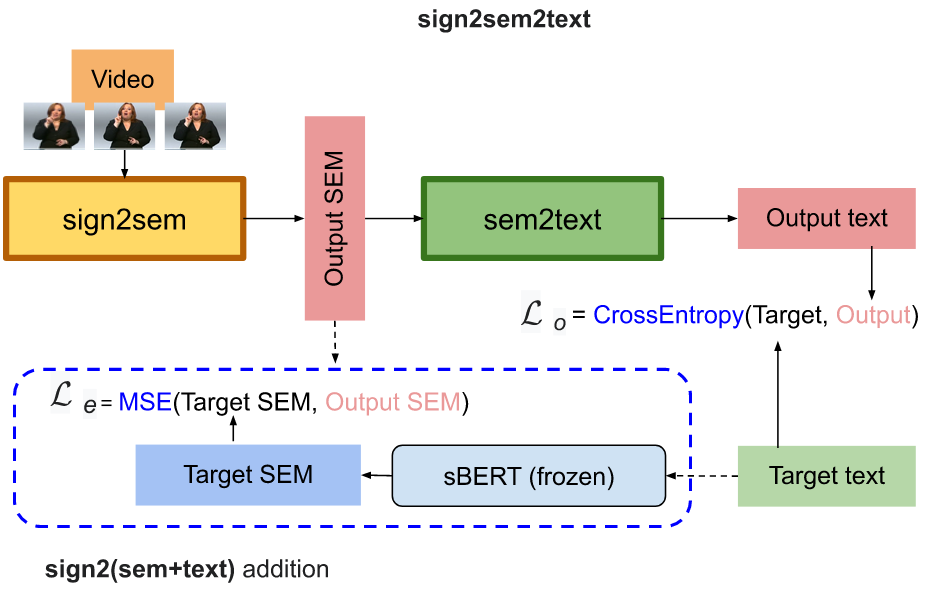}
    \caption{sign2sem and sem2text independent modules for the SLT task (left plot).  End2end architectures: pipeline system sign2sem2text and multitask system sign2(sem+text) (right plot).}
    \label{fig:archs}
\end{figure*}

\section{Related Work}
\label{s:sota}

\citet{camgoz18} proposed three formalisations considering SLT as a seq2seq problem that converts a sequence of signs into a sequence of words: \Ni sign2text, a model that encodes video frames using pretrained 2D CNNs as spatial features and then uses an RNN to generate the text; \Nii gloss2text, a model learning the translation between a sequence of textual glosses and fluent spoken language text; and \Niii a sign2gloss2text model that adds an extra intermediate gloss layer between the video and output text levels of a  sign2text architecture to provide additional gloss supervision using a CTC loss. 


In follow-up work, \citet{camgoz2020} proposed an architecture for joint learning continuous sign language recognition (CSLR) and SLT which uses the same input as \citet{camgoz18, csldaily}, that is, pretrained visual features, but a transformer~\cite{Vaswani:2017} for text generation. \citet{camgoz2020} conjectured that gloss2text results with ground-truth glosses provide an upper bound for SLT.
Supporting this assumption, their translation quality on PHOENIX-2014T as measured by BLEU \cite{papineni-etal-2002-bleu} achieved 24.5 on gloss2text and 21.8 on sign2text.

\citet{yin-read-2020-better} used a different visual representation with a multi-cue network \cite{zhou2020spatial} to encode videos. Cues included face, hands and pose besides the full frame.
With a BLEU score of 24.0 they impoved over sign2gloss2text \citet{camgoz2020} 
\hy{and} concluded that their visual representation was better than the spatial frame  embeddings used by the \citet{camgoz2020}. 
\citet{chenEtAl:2022,chen2022two} used both pretraining of a network based on S3Ds \citep{xieEtAl:2018} on action recognition for CSLR (sign2gloss) and pretraining of a textual transformer (gloss2text) with mBART-25 \citep{liu-etal-2020-multilingual-denoising}. Both types of pretraining are progressively adapted to the domain of the task by adding data closer to the domain. An additional mapping network between the vision and language parts allows \citet{chenEtAl:2022} to build an end2end sign2text model relying on internal gloss supervision. To the best of our knowledge, \citet{chen2022two} is the current state of the art for both sign2text (BLEU=28.95) and sign2gloss2text (BLEU=26.71), \hy{all on the PHOENIX-2014T data set}.

Over the last few years, several gloss-free models have emerged
\citep{gfTspnet20, gfCagcr22, gfGsgslt23, lin-etal-2023-gloss}. \citet{Zhou_2023_ICCV} obtains the current state-of-the-art in this category by utilising visual-language pretraining following CLIP \citep{radford2021learning}. 
On the datasets \hy{(\citet{camgoz18, csldaily})} where the two approaches can be compared, translation quality diminishes by up to 7 BLEU points when the glosses are not used  \citep{gfGsgslt23, Zhou_2023_ICCV}. 
\citet{Tarres_2023_CVPR} uses the How2Sign dataset (\citet{duarteEtAl:2021}) (where no gloss information is available) with I3D (\citet{i3dJcarreira}) features  for video representations and a Transformer.
\citet{UthusSLhw1} introduces a new dataset, YouTube-ASL, 10 times larger than the previous one (\citet{duarteEtAl:2021}), and uses 2D pose estimation and pretraining to improve
on \citet{Tarres_2023_CVPR} best results on How2Sign (BLEU=8.09 vs BLEU=12.4). 
Simultaneously with \hy{our work},
\citet{rust2024privacyaware} pretrains a self-supervised and privacy-aware visual model on YouTube-ASL to achieve the new state-of-the-art performance on How2Sign (BLEU=15.5).

\section{SEM-based Architectures}
\label{s:arch}
In our work we build two systems that revolve around textual sentence embeddings, SEM, as depicted in Figure~\ref{fig:archs}. The figure presents two independent modules sign2sem and sem2text (left plots) that we later combine in sign2sem2text and sign2(sem+text) in an end2end setting (right plot). 

\paragraph{$\bullet$ sign2sem Module}
This module predicts an intermediate SEM vector. Given a set of frames (video) features, sign2sem produces a vector representing the sentence signed in the video using a transformer encoder. 
\paragraph{~~Pretraining} the visual feature sentence embedding model on text.
We follow \citet{reimers-gurevych-2019-sentence} and train a Siamese network with twin subnetworks 1 and 2. We compute the loss as the minimum squared error (MSE):
\begin{equation*}
\mathcal{L}_e = \frac{1}{N} \sum^{N}_{i=1}\left(\cos(S_{1,i}, S_{2,i}) - \cos(E_{1,i}, E_{2,i})\right)^2
\label{eq:le}
\end{equation*} where $N$ is the batch size, and $S$ and $E$ contain the target text SEM vectors and the predicted output SEM vectors respectively. 
In our experiments, the target SEM vector is given by sBERT \citep{reimers-gurevych-2019-sentence} here and in our models below.

\paragraph{$\bullet$ sem2text Module} This module is responsible for the text reconstruction from sentence embeddings SEM. It produces the text translation of the video features encoded in a given SEM vector. The core sem2text model is a transformer model; we compare encoder--decoder and only decoder systems for the task:

\textbf{- Encoder--decoder (SLTr)}: this version uses the sign language transformer (SLTr) architecture as in \citet{camgoz2020}. We use a transformer base with a linear projection from the SEM vector input instead of the usual word embedding layer. 

\textbf{- Decoder only with pretrained mBART}: this version uses a pre-trained mBART-25 decoder and a linear layer to project the SEM vectors into the mBART model dimensions.

\paragraph{~~Pretraining}
We train both transformers (SLTr from scratch and the already pretrained mBART-25) with Wikipedia data and then finetune them on the SL datasets.
We compute the translation output loss as the cross-entropy:

\vspace{-1.5em}
\begin{equation*}
\mathcal{L}_o = {\rm CE}(T,O) =  -\frac{1}{N} \sum_{i=1}^{N} \sum_{j=1}^{M} \left(\text{T}_{ij} \cdot \log(\text{O}_{ij})\right) 
\label{eq:lo}
\end{equation*}

\vspace{-0.5em}
\noindent
where $N$ is the batch size, \(M\) the vocabulary size, $T$ is the target text and $O$ is the output text.


\paragraph{}
After pretraining each component (sign2sem and sem2text), we combine them together for end2end training. We explore two approaches: an approach that only uses the output loss \(\mathcal{L}_o\), \textbf{sign2sem2text}, and an approach that integrates an additional supervision loss \(\mathcal{L}_e\), \textbf{sign2(sem+text)}.

\begin{table*}[htp]
 \centering
 \resizebox{\textwidth}{!}{%
\begin{tabular}{ll cccc crcc}
\toprule
 && \mc{4}{c}{PHOENIX-2014T (DGS)} & \mc{4}{c}{How2Sign (ASL)}\\
 \cmidrule(lr){3-6}
 \cmidrule(lr){7-10}
 && \mc{1}{c}{BLEU$_{\rm val}$} & \mc{1}{c}{BLEU} & \mc{1}{c}{chrF} & \mc{1}{c}{BLEURT} & \mc{1}{c}{BLEU$_{\rm val}$} & \mc{1}{c}{BLEU} & \mc{1}{c}{chrF} & \mc{1}{c}{BLEURT}\\
 \midrule
 \multirow{4}{*}{\rotatebox[origin=c]{90}{SLTr}}
&sign2sem2text - mono    & 14.22  & 13.4$\pm$1.4 & 33.5$\pm$1.5 & 0.379$\pm$0.016 & ~6.69 & 5.7$\pm$0.4 & 21.2$\pm$0.4 & 0.382$\pm$0.005\\
&sign2sem2text - multi  & 13.05 & 12.7$\pm$1.3 & 32.3$\pm$1.3 & 0.343$\pm$0.014 & ~6.48 & 6.4$\pm$0.4 & 22.0$\pm$0.5 & 0.403$\pm$0.006\\
&sign2(sem+text) - mono  & 19.10 & 18.8$\pm$1.7 & 40.1$\pm$1.5 & 0.437$\pm$0.016 & 10.41 & 9.5$\pm$0.5 & 27.4$\pm$0.5 & 0.445$\pm$0.006\\
&sign2(sem+text) - multi  & 17.03 & 16.6$\pm$1.6 & 37.9$\pm$1.5 & 0.412$\pm$0.016 & ~7.85 & 7.8$\pm$0.4 & 25.4$\pm$0.5 & 0.430$\pm$0.006 \\
\midrule
 \multirow{4}{*}{\rotatebox[origin=c]{90}{mBART}}
&sign2sem2text - mono  & 16.67 & 17.3$\pm$1.6 & 38.2$\pm$1.5 & 0.434$\pm$0.016 & ~9.32 &  9.8$\pm$0.5 & {\bf 31.2$\pm$0.5} & 0.477$\pm$0.006\\
&sign2sem2text - multi  & 16.91 & 16.5$\pm$1.6 & 37.3$\pm$1.5 & 0.425$\pm$0.016 & ~9.11 & 9.6$\pm$0.5 & {\bf 31.2$\pm$0.5} & 0.475$\pm$0.006\\
&sign2(sem+text) - mono & 24.07 & {\bf 24.2$\pm$1.9} & {\bf 46.3$\pm$1.6} & {\bf 0.483$\pm$0.017} & 12.20 & {\bf 11.7$\pm$0.5} & {\bf 32.0$\pm$0.5} &  {\bf 0.487$\pm$0.006}\\
&sign2(sem+text) - multi  & 24.12 & {\bf 24.1$\pm$1.9} & {\bf 46.1$\pm$1.6} & {\bf 0.481$\pm$0.017} & 12.34 & {\bf 12.0$\pm$0.5} & {\bf 31.8$\pm$0.5} & {\bf 0.483$\pm$0.006} \\
\bottomrule
 \end{tabular}
}
 \caption{Translation performance of our models on validation (val) and test. Best models at 95\% confidence level are highlighted. Previous state-of-the-art for gloss-free systems is BLEU=21.44 for PHOENIX \citep{Zhou_2023_ICCV} and 8.03 for How2Sign~\citep{Tarres_2023_CVPR}. \citet{chen2022two} achieves 28.95 on PHOENIX with their gloss-assisted system sign2text and 26.71 with sign2gloss2text. \citet{rust2024privacyaware} achieves 15.5 on How2Sign pretraining with YouTube-ASL.}
 \label{tab:fullSystem}
\end{table*}

\paragraph{$\bullet$ sign2sem2text}
is a simple pipeline combination of sign2sem and sem2text where the output SEM of the first module is used as input by the second module to obtain the final text prediction. The two pretrained modules (with both variants {\bf SLTr} and {\bf mBART}) are put together and trained in an end2end manner without any intermediate supervision. This formalisation is the sentence embedding equivalent to the  sign2gloss2text approach.

\paragraph{$\bullet$ sign2(sem+text)}
performs translation using the same components as sign2sem2text. However, it uses the sign2sem SEM output as additional intermediate supervision using MSE loss computed against the target text SEM in a multitask learning approach. Both, \(\mathcal{L}_e\) (sentence embedding) and \(\mathcal{L}_o\) (output text), are used jointly to train the model.

For {\bf SLTr}, we take the SEM before the tanh and pooling (see Figure~\ref{fig:archs} (left--middle)), and project it into the SLTr model dimension. The supervision is applied after the SLTr encoder. For {\bf mBART}, the supervision happens right before the mBART.

\paragraph{}Our architectures can be trained both monolingually and mutilingually simply by using multilingual embeddings and merging multilingual training data.


\section{Experimental Settings}
\label{s:settings}

We use two diverse (language and domain) {\bf datasets} for our experiments:
\paragraph{RWTH-PHOENIX-2014T \cite{camgoz18}} 11 hours of weather forecast videos from 9 signers. Signers use German Sign Language and both transcriptions and glosses are available.
\vspace{-0.5em}
\paragraph{How2Sign \cite{duarteEtAl:2021}} 80 hours of instructional videos with speech and transcriptions and their corresponding American Sign Language videos (glosses unavailable) from 11 signers.

\medskip
Detailed statistics for each dataset are provided in Appendix~\ref{app:data}. We {\bf preprocess} the textual part of the datasets in a way that allows us to compare to the results obtained by \citet{camgoz18}. We tokenise and lowercase the input for both training and evaluation. 
We apply BPE \citep{sennrich-etal-2016-neural} with a vocabulary size of 1500 for Phoenix-2014T and 5000 for How2Sign. When pretraining sem2text SLTr, we use a shared (en--de) vocabulary size of 32000. In cases where we use pretrained models, we keep the tokenisation of the model.

For video files, we extract frames using ffmpeg. We normalise the images, and resize them to 224x224.
In this step, we initially obtain frame features from a pretrained model \citep{pmlr-v97-tan19a}, which does not contain gloss information. We then apply pooling to remove the spatial dimensions, followed by batch normalisation with ReLU, following the approach outlined by \citet{camgoz2020}.
This generic approach facilitates the combination of datasets in the multilingual setting.

We use two multilingual {\bf pretrained models} that cover both German and English, sBERT~\citep{reimers-gurevych-2019-sentence}%
\footnote{We use all-MiniLM-L12-v2 model with 384 dimensions.}
for sentence embeddings and mBART~\citep{liu-etal-2020-multilingual-denoising}%
\footnote{We use mBART-25 1024 dimensions.}
 as a language model. For further pretraining we use 26 million sentences per language from the English and German Wikipedia dumps extracted with Wikitailor \citep{EspanaBonetEtal:2022wikiTailor}.

\medskip
Following~\citet{muller-etal-2022-findings} and \citet{muller-etal-2023-findings}, we {\bf evaluate} the models using three common automatic metrics in machine translation: BLEU~\cite{papineni-etal-2002-bleu}, chrF~\cite{popovic-2015-chrf} and BLEURT~\cite{sellam-etal-2020-bleurt}. Specifics can be found in Appendix~\ref{app:eval}.
In all cases, we estimate 95\% confidence intervals (CI) via bootstrap resampling~\cite{koehn-2004-statistical} with 1000 samples.

\section{Results and Discussion}
\label{s:results}

Table \ref{tab:fullSystem} presents the results for our models and variants. Two major trends are observed: \Ni massive pretraining of the sem2text module (mBART vs SLTr) significantly improves the results, confirming the observations by \citet{chenEtAl:2022} and \Nii the multitask approach sign2(sem+text) is better than the pipeline approach sign2sem2text. These findings hold for all three evaluation metrics at 95\% confidence level.



Potentially beneficial effects of multilinguality are less evident. Monolingual and multilingual approaches are not distinguishable within the 95\% CIs, possibly due to  large differences in the domain of the datasets preventing effective transfer between languages.


Our best system, sign2(sem+text) with the pretrained text decoder, achieves state-of-the-art results on How2Sign when no additional SLT dataset is used for pretraining, improving from 8 to 12 BLEU points over \citet{Tarres_2023_CVPR}. For PHOENIX-2014T, we surpass all previous gloss-free approaches (24 vs 21 BLEU), but we are still below the best approach that uses glosses \citep{chen2022two} (24 vs 29 BLEU).

\begin{figure}[ht]
        \includegraphics[width=0.45\textwidth]{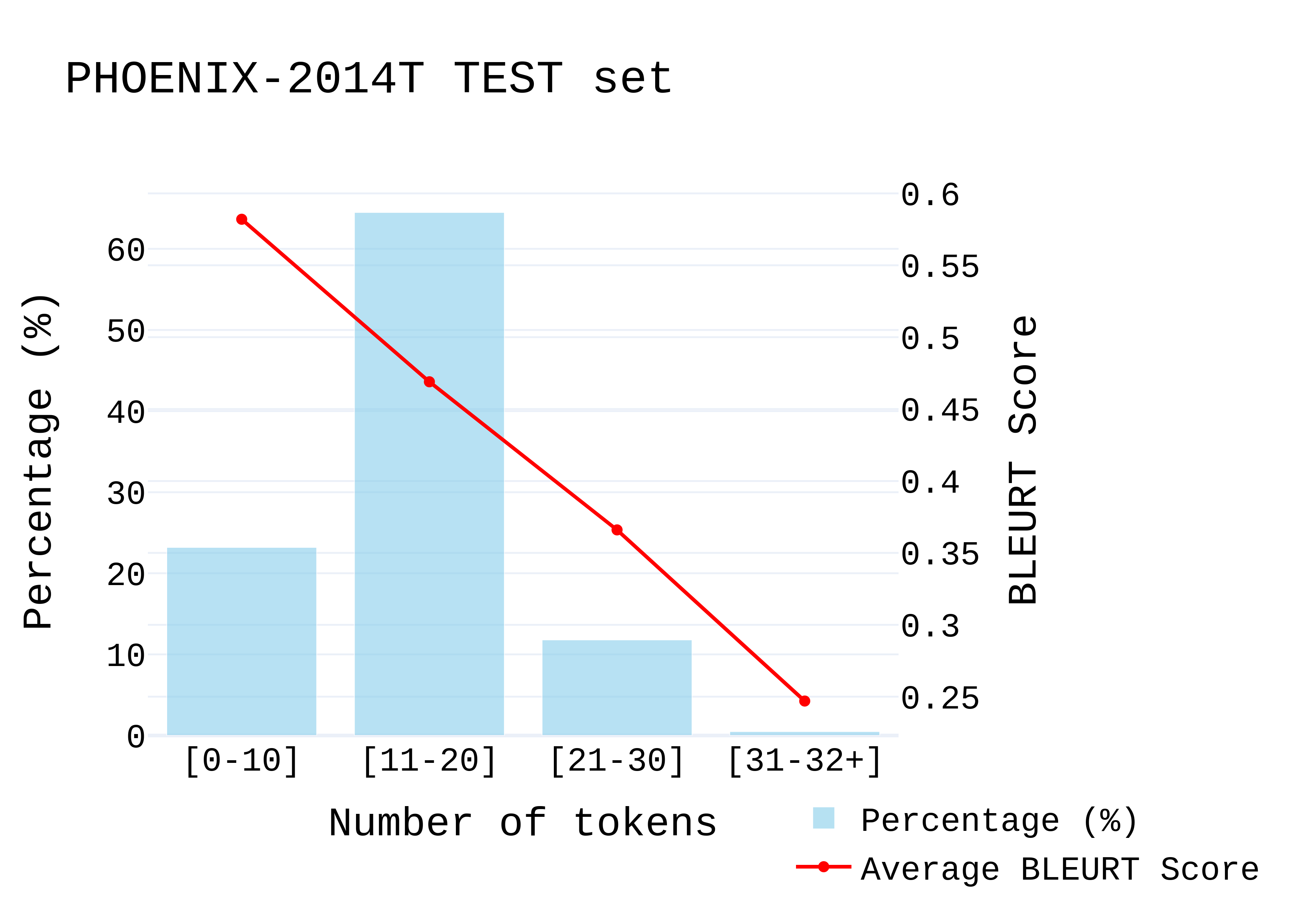}
       \hfill
        \includegraphics[width=0.45\textwidth]{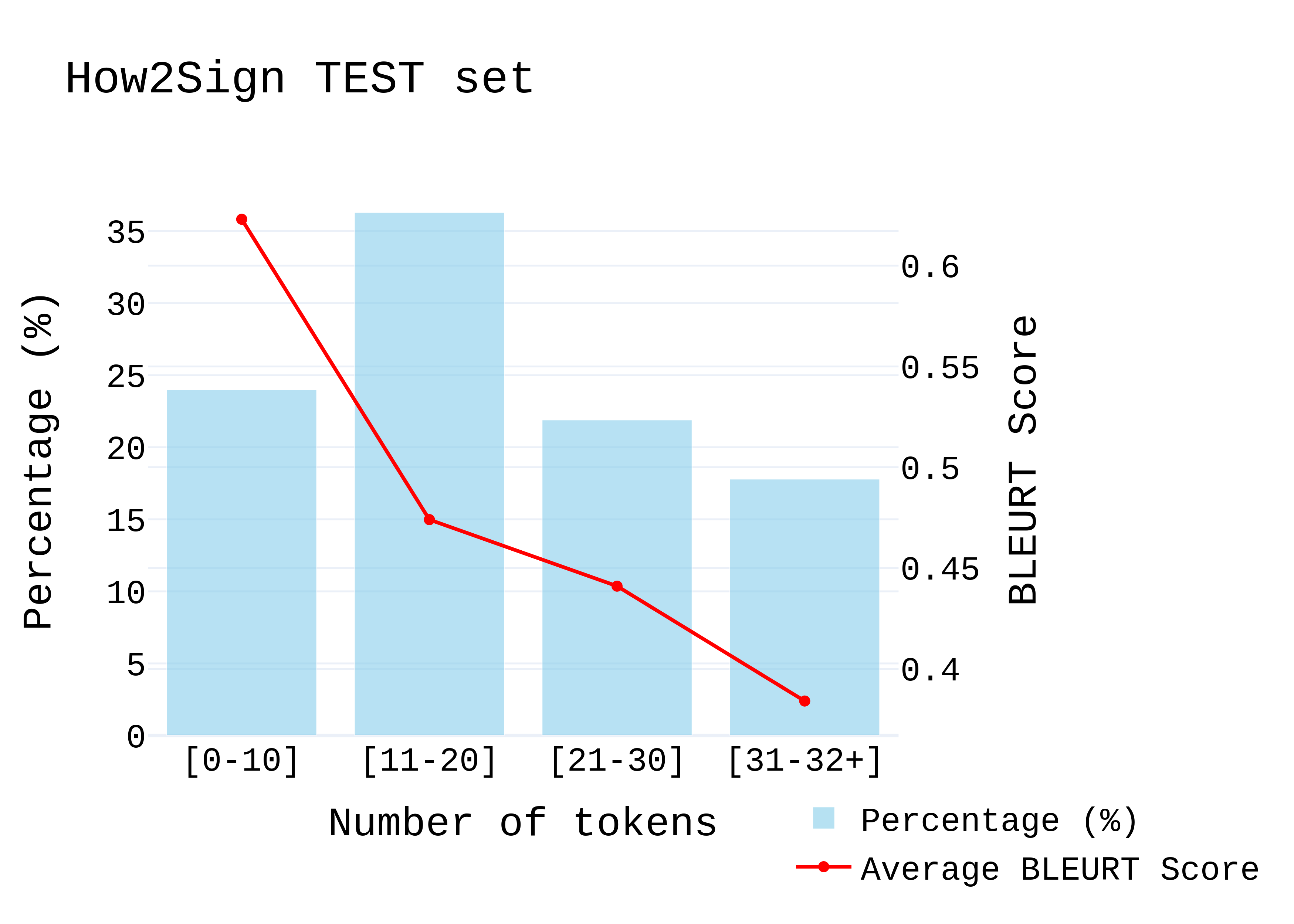}
    \caption{Average BLEURT score on different token length intervals on PHOENIX-2014T and How2Sign test.}
    \label{fig:toklen}
\end{figure}

\begin{table}[t]
 \centering
 \resizebox{1\columnwidth}{!}{%
\begin{tabular}{l cccc}
 & \mc{4}{c}{PHOENIX-2014T (DGS)} \\
 \cmidrule(lr){2-5}
 & \mc{1}{c}{BLEU$_{\rm val}$} & \mc{1}{c}{BLEU} & \mc{1}{c}{chrF} & \mc{1}{c}{BLEURT} \\
 \midrule
Camg\"oz (2020)     & 20.16 & 19.26 & -- & -- \\
Chen (2022a)  & 27.61 & 28.39 & -- & -- \\
 \midrule
SLTr - mono    & 31.53 & 30.1$\pm$2.0& 52.2$\pm$1.7 & 0.526$\pm$0.018\\
SLTr - multi   & 29.20 & 31.3$\pm$2.1 & 52.9$\pm$1.7 & 0.530$\pm$0.018\\
\midrule
mBART - mono    & 37.11 & {\bf 38.0$\pm$2.4} & {\bf 57.5$\pm$1.9} & {\bf 0.588$\pm$0.019} \\
mBART - multi   & 36.91 & {\bf 37.5$\pm$2.3} & {\bf 57.4$\pm$1.8} & {\bf 0.584$\pm$0.018} \\
\midrule
\\
~\\
 & \mc{4}{c}{How2Sign (ASL)}\\
 \cmidrule(lr){2-5}
 & \mc{1}{c}{BLEU$_{\rm val}$} & \mc{1}{c}{BLEU} & \mc{1}{c}{chrF} & \mc{1}{c}{BLEURT}\\
 \midrule
SLTr  - mono    & 13.24  & 14.6$\pm$0.6 & 34.3$\pm$0.6 & 0.489$\pm$0.006\\
SLTr  - multi   & 16.17 & 16.4$\pm$0.7 & 36.5$\pm$0.7 & 0.529$\pm$0.007\\
\midrule
mBART  - mono   & 23.04 & {\bf 22.8$\pm$1.0} & {\bf 43.3$\pm$0.9} & {\bf 0.577$\pm$0.008}\\
mBART  - multi  & 24.60 & {\bf 23.3$\pm$1.0} & {\bf 43.9$\pm$0.9} & {\bf 0.571$\pm$0.008}\\
\bottomrule
 \end{tabular}
}
 \caption{Reconstruction quality for the sem2text subtask of our models and gloss2text state-of-the-art on validation (val) and test. Best models at 95\% confidence level are highlighted. }
 \label{tab:semSystem}
\end{table}

\paragraph{Reconstruction quality: sem2text.}
In our approach, sentence embeddings take the role of manually produced glosses in previous work. Our sem2text translation module defines the upper-bound results for the full system as gloss2text did in previous work.
Our best sign2(sem+text) models with mBART produce a reconstruction score of BLEU 38.0$\pm$2.4/23.3$\pm$1.0, chrF 57.5$\pm$1.9/43.9$\pm$0.9 and BLEURT 0.588$\pm$0.019/0.571$\pm$0.008 for PHOENIX-2014T/How2Sign (see Table~\ref{tab:semSystem}). Where the comparison with glosses is available (PHOENIX), we improve over gloss2text by up to 10 BLEU points. 
We hypothesise that a sentence is better represented by its embedding than by a string of glosses and this explains why the translation quality for sem2text is higher than for gloss2text.
If these components (sem2text and gloss2text) are the upper-bound to the end2end sign2text translation, SEM-based systems are potentially at an advantage.
These results, together with the fact that SEM models can be applied to raw data without annotations, highlight the promising future prospects of, especially, sign2(sem+text).

\paragraph{SEM-based vs gloss-based SLT.}
For comparison purposes, we integrate SEM supervision in a state-of-the-art gloss-based SLT system, Signjoey \citep{camgoz2020}, by replacing their gloss supervision by SEM supervision. We perform no pretraining and train the two systems under the same conditions. We observe that convergence with SEM is faster and requires less than half of the iterations to finish (5k vs 12k) using the same setting and resources. The detailed training evolution is shown in Appendix~\ref{app:joey}.

\paragraph{Translation quality vs output length.}
Figure \ref{fig:toklen} shows the token length distribution of PHOENIX-2014T and How2Sign along with the average BLEURT score on each interval. The equivalent plots for chrF and BLEU are in Figures \ref{fig:chbl} and \ref{fig:chbl1} in Appendix \ref{app:abb} respectively. In the PHOENIX test set, almost 90\% of the sentences contain 20 tokens or less, while the number decreases to 60\% for How2Sign. The 10-20 token range is the one with the best scores. While the drop in performance in translation quality for long sentences is smaller in How2Sign, the difference in the distribution affects the global quality.

\section{Conclusions}
\label{s:conclusions}

We present a new approach to sign language  translation using automatically computed sentence embeddings instead of manual gloss labels as intermediate representation with (sign2(sem+text)) and without (sign2sem2text) SEM supervision. 
We outperform the state-of-the-art of gloss-free SLT
when no additional SLT datasets are used for
pretraining\hy{,} closing the gap to gloss-based SLT. 

According to the upper-bound set by sem2text translation quality, there is still room for improvement for the end2end SEM-based SLT models. In this work, we limited ourselves to existing visual feature extractors, in the future we plan to train a SEM-based visual feature extractor on SL datasets in order to get closer to our sem2text upper-bound and match gloss-based performance.

\section*{Limitations}
\label{s:limitations}
Our SL datasets cover American English and German. Sentence embeddings for these languages are good quality as lots of textual data is available for pre-training. It remains to be studied how the quality of the embeddings affects the final translation quality. This is important for low-resourced languages, i.e. languages with limited amounts of monolingual text data but, to the best of our knowledge, no public sign language data set exists for them.


\section*{Acknowledgements}
This work has been funded by 
by BMBF (German Federal Ministry of Education
and Research) via the project SocialWear (grant
no. 01IW20002).

\bibliography{custom}
\appendix

\clearpage
\section{Datasets Statistics}
\label{app:data}

Table~\ref{tab:copora} summarises the statistics for the corpora used in the experiments.


\begin{table}[ht]
\centering
\small
\begin{tabular}{l r r}
\toprule
  & Phoenix-2014T & How2Sign \\
  \midrule
  Src. Lang. & German & Am. English \\
  Tgt. Lang & DGS & ASL \\
  Hours & 11 & 80 \\
  Signers & 9 & 11 \\
  Sentences & 7000 & 35191\\
  Val. Size & 540 & 1741 \\
  Test Size & 629 & 2322 \\ 
\bottomrule
 \end{tabular}
 \caption{Statistics of the corpora used in the experiments. Source (Src.Lang.) and target (Tgt.Lang.) refer to the direction in which the corpora were created; all our experiments involve sign2text.} 
 \label{tab:copora}
\end{table}

\section{Infrastructure and Network Hyperparameters}
\label{app:parameters}

We implement our SLT framework using PyTorch, and libraries from sBERT \citep{reimers-gurevych-2019-sentence} and Huggingface \citep{wolf2019huggingface}. Our code is publicly available at Github.%
\footnote{\url{https://github.com/yhamidullah/sem-slt}}

Tables~\ref{tab:sign2sem}, \ref{tab:SLTr} and \ref{tab:mBART} show the hyperparameters and training times for the sign2sem and sem2text with SLTr and mBART transformers respectively. We run our experiments using 8 A100-80GB GPUs. For sign2sem2text and sign2(sem+text), each experiment runs for 72 hours and the configurations are inherited from the standalone modules sign2sem and sem2text.

\begin{table}[ht]
\centering
\small
\begin{tabular}{cc}
\hline
\multicolumn{1}{|c|}{\textbf{Parameter}}       & \multicolumn{1}{c|}{\textbf{Value}}    \\ \hline
\multicolumn{1}{|c|}{model}                    & \multicolumn{1}{c|}{all-MiniLM-L12-v2} \\ \hline
\multicolumn{1}{|c|}{batch\_size\_per\_device} & \multicolumn{1}{c|}{16}                \\ \hline
\multicolumn{1}{|c|}{learning\_rate}           & \multicolumn{1}{c|}{1e-5}              \\ \hline
\multicolumn{1}{|c|}{input\_projection\_dim}   & \multicolumn{1}{c|}{1024}              \\ \hline
\multicolumn{1}{|c|}{scheduler}  & \multicolumn{1}{c|}{warmuplinear}   \\ \hline                       
\multicolumn{1}{|c|}{\textbf{Training time}}   & \multicolumn{1}{c|}{72 hours (5 GPU)} \\ \hline
\end{tabular}
\caption{Hyperparameters for the sign2sem module, we use the defaults of sBERT trainer for the rest.}
    \label{tab:sign2sem}
\end{table}

\begin{table}[ht]
\centering
\small
\begin{tabular}{|c|c|}
\hline
\textbf{Parameter}              & \textbf{Value}    \\ \hline
num\_encoder\_layers            & 3                 \\ \hline
num\_decoder\_layers            & 3                 \\ \hline
d\_model                        & 512               \\ \hline
ff\_size                        & 2048              \\ \hline
input\_projection\_dim          & 1024              \\ \hline
batch\_size\_per\_device\_train & 32                \\ \hline
batch\_size\_per\_device\_val   & 32                \\ \hline
learning\_rate                  & 1e-5              \\ \hline
lr\_scheduler                   & reduceLROnPlateau \\ \hline
freeze\_word\_embeddings        & True              \\ \hline
\textbf{Training time}          & 1 hour (1GPU) \\ \hline
\end{tabular}
\caption{Hyperparameters for the sem2text module with SLTr transformer, the rest are inherited from \citet{camgoz2020}.}
    \label{tab:SLTr}
\end{table}

\begin{table}[ht]
\centering
\small
\begin{tabular}{|c|c|}
\hline
\textbf{Parameter}              & \textbf{Value} \\ \hline
input\_projection\_dim          & 1024           \\ \hline
batch\_size\_per\_device\_train & 4              \\ \hline
batch\_size\_per\_device\_val   & 4              \\ \hline
learning\_rate                  & 1e-5           \\ \hline
fp16                            & True           \\ \hline
freeze\_word\_embeddings        & True           \\ \hline
\textbf{Training time}          & 156 hours (8 GPUs) \\ \hline
\end{tabular}
\caption{Hyperparameters for the sem2text module with mBART decoder, the rest are inherited from the Huggingface trainer default values.}
    \label{tab:mBART}
\end{table}

\section{Automatic Evaluation}
\label{app:eval}

Following~\citet{muller-etal-2022-findings} and \citet{muller-etal-2023-findings}, we evaluate the models using three common automatic metrics in machine translation: BLEU~\cite{papineni-etal-2002-bleu}, chrF~\cite{popovic-2015-chrf} and BLEURT~\cite{sellam-etal-2020-bleurt}.
Notice that even though other semantic metrics based on embeddings might correlate better with human judgements \citep{kocmi-etal-2021-ship,freitag-EtAl:2022:WMT}, they cannot be used for sign language translation because the source is video and not text.
We use sacreBLEU~\cite{post-2018-call} for BLEU%
\footnote{\texttt{BLEU|nrefs:1|bs:1000|seed:16|case: mixed|eff:no|tok:13a|smooth:exp|version: 2.4.0}}
and chrF%
\footnote{\texttt{chrF2|nrefs:1|bs:1000|seed:16|case: mixed|eff:yes|nc:6|nw:0|space:no|version: 2.4.0}}
and the python library for BLEURT.%
\footnote{BLEURT v0.0.2 using checkpoint BLEURT-20.}

Previous work starting with \citet{camgoz18} does mainly report only BLEU scores, but they do not specify the BLEU variant used or the signature in sacreBLEU. Therefore, comparisons among systems might not be strictly fair.

\section{Gloss-based vs SEM-based Systems' Training Performance}
\label{app:joey}

\begin{figure}[t]
\centering
        \includegraphics[width=0.45\textwidth]{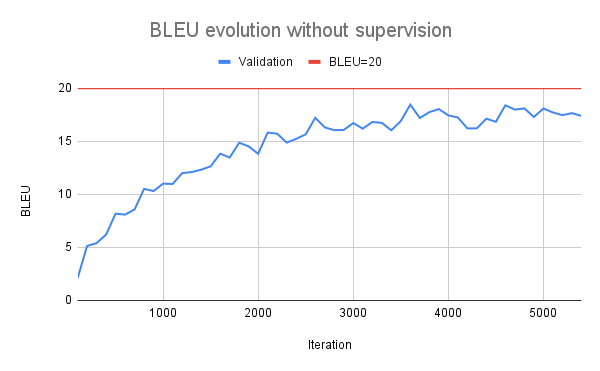}
       \hfill
        \includegraphics[width=0.45\textwidth]{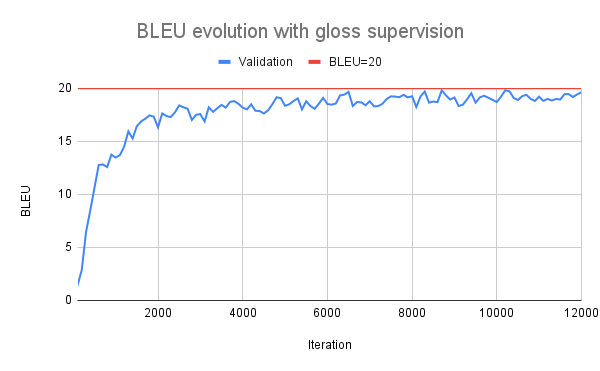}
        \hfill
        \includegraphics[width=0.45\textwidth]{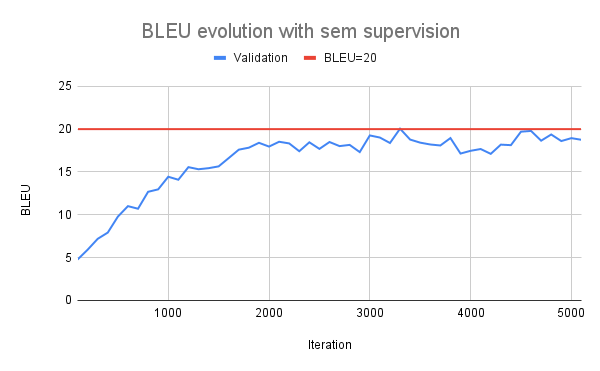}
    \caption{Validation BLEU on PHOENIX without supervision (top plot), with gloss supervision (middle plot) and with SEM supervision (bottom plot).}
    \label{fig:joeycomp}
\end{figure}

Figure~\ref{fig:joeycomp} shows the training evolution for a simple SLT system with no additional supervision (top), additional gloss supervision (middle) and SEM supervision (bottom) implemented in the Signjoey framework~\citep{camgoz2020} and trained on PHOENIX-2014T. We use the best hypeparameters in \citet{camgoz2020} and add our SEM supervision as a replacement of their recognition loss. 

The three plots in Figure~\ref{fig:joeycomp} include a red line at translation quality BLEU=20 for reference. The first thing to notice is that both supervision methods reach the red line, but the one lacking any additional supervision lays behind. Second, we observe that the system with the additional SEM supervision reaches BLEU=20 earlier than the system with glosses: the gloss system needs 12k to finish and only 5k iterations are needed in the case of SEM. In both cases, we use early stopping with BLEU  patience 7.
Finally, notice that the gloss and SEM systems achieve the same translation quality but one does not need any data annotation with SEM.


\section{Ablation Study on Sentence Length}
\label{app:abb}
Following the analysis of Section~\ref{s:results}, we include the translation quality scores BLEU and chrF per sentence length.


\begin{figure}[t]
\centering
        \includegraphics[width=0.45\textwidth]{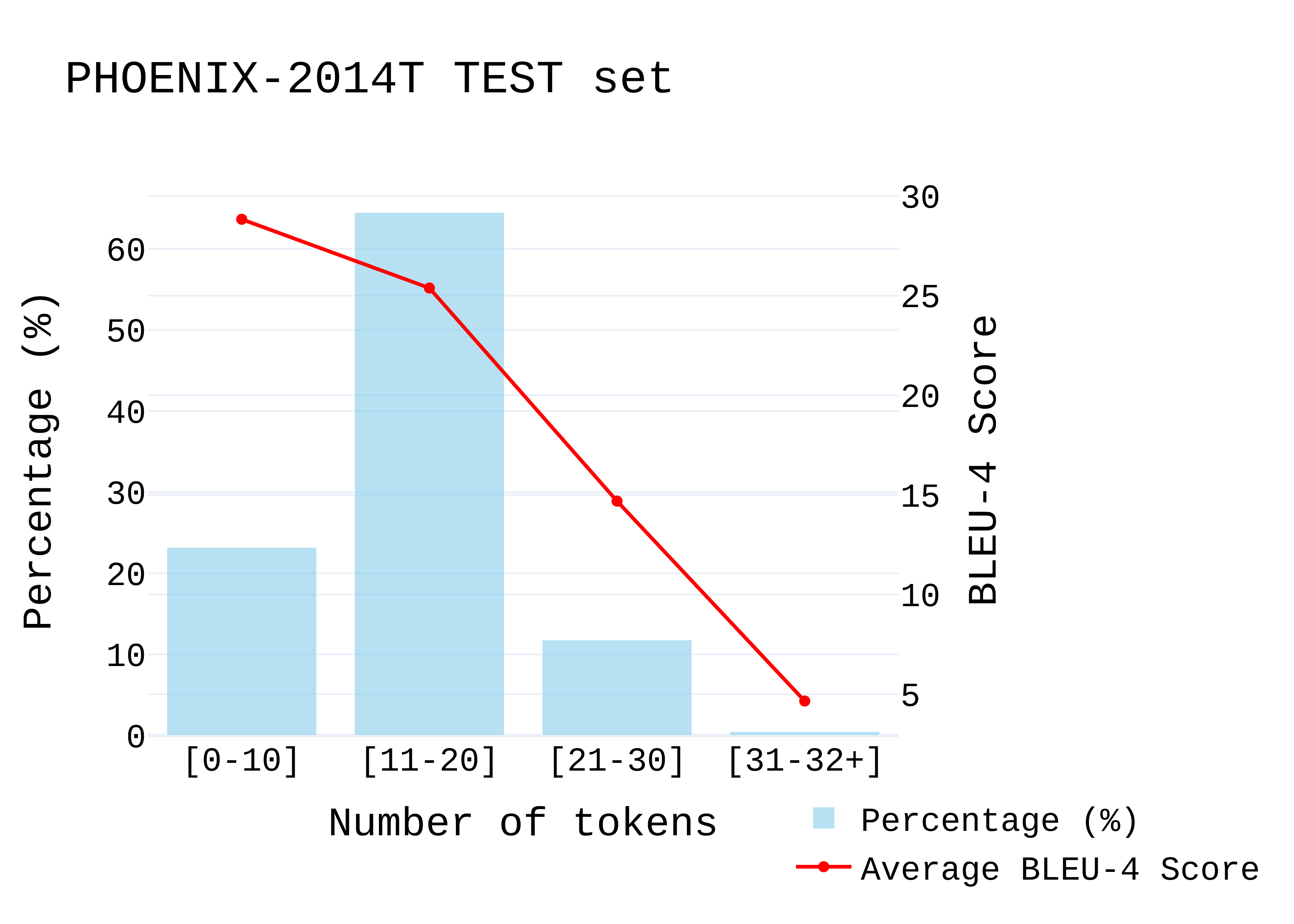}
       \hfill
        \includegraphics[width=0.45\textwidth]{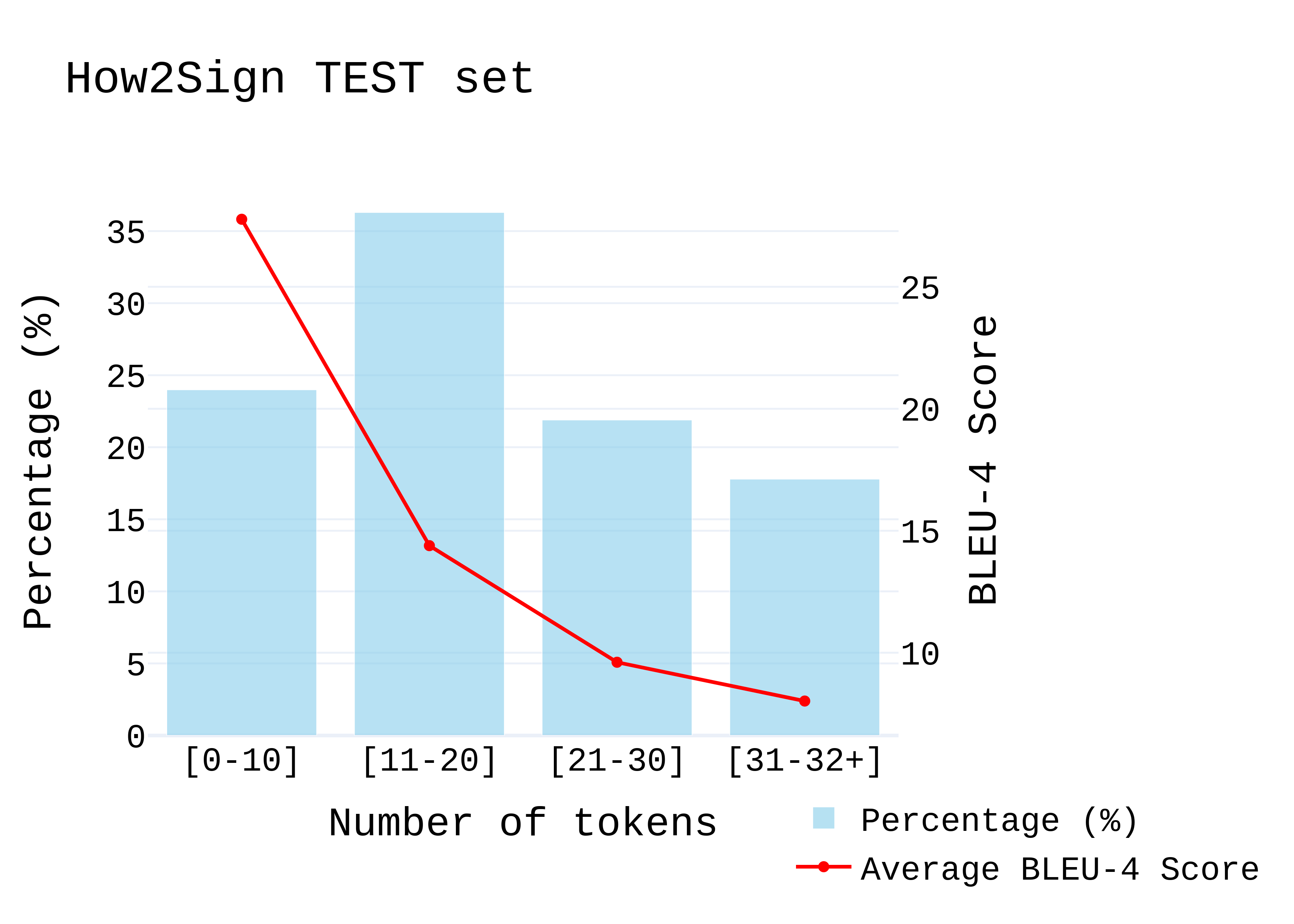}
    \caption{Variation of the average BLEU
    score on different token length intervals on PHOENIX-2014T (top) and How2Sign (bottom) test sets.}
    \label{fig:chbl}
\end{figure}

\begin{figure}[t]
\centering
        \includegraphics[width=0.45\textwidth]{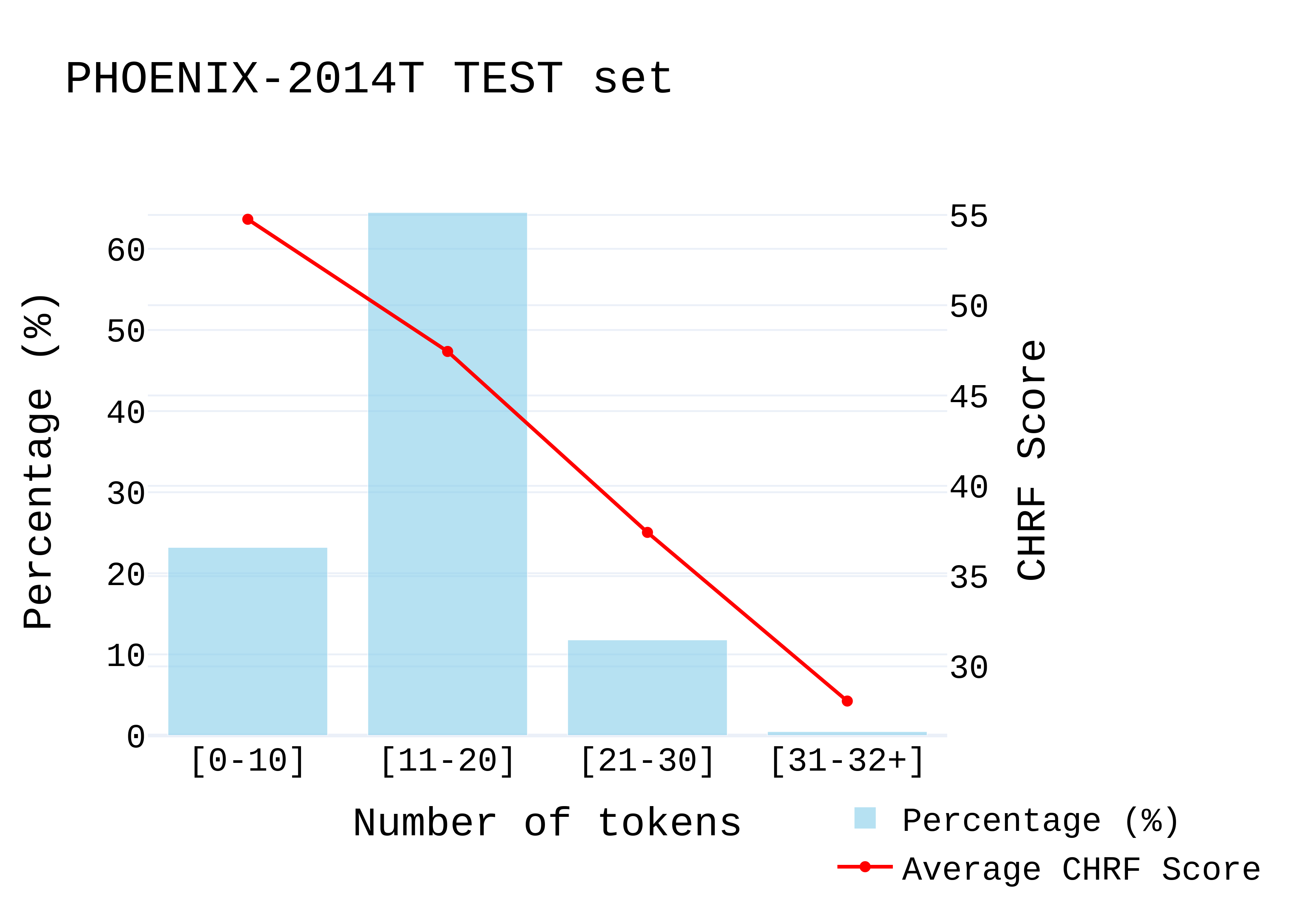}
        \hfill
        \includegraphics[width=0.45\textwidth]{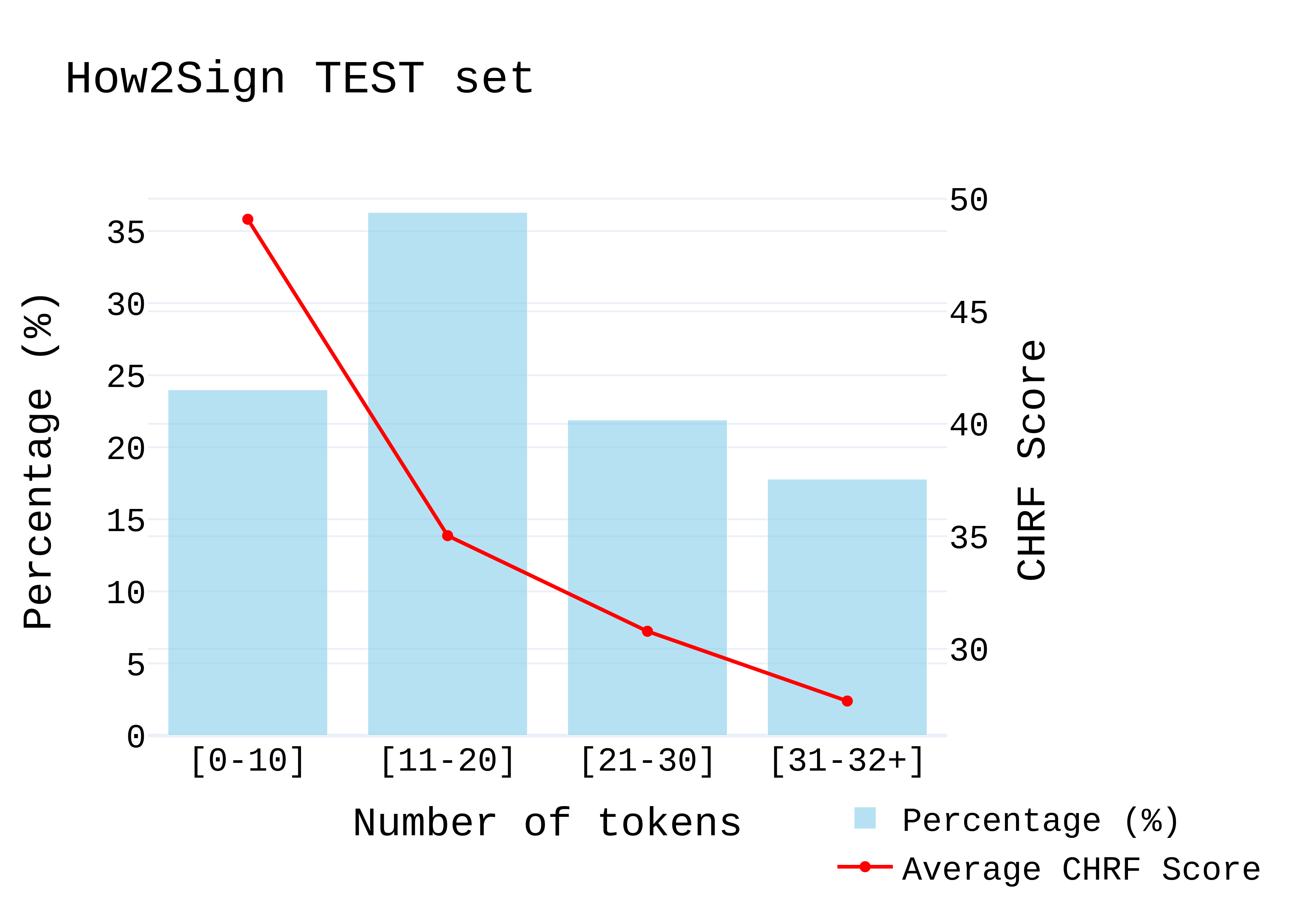}
    \caption{Variation of the average 
    chrF score on different token length intervals on PHOENIX-2014T (top) and How2Sign (bottom) test sets.}
    \label{fig:chbl1}
\end{figure}

\end{document}